\documentclass[letterpaper]{article}
\usepackage{aaai2026} 
\nocopyright
\usepackage{times} 
\usepackage{helvet} 
\usepackage{courier} 
\usepackage[hyphens]{url} 
\usepackage{graphicx} 
\usepackage{graphicx} 
\usepackage{natbib} 
\usepackage{caption} 
\usepackage{amsmath}
\usepackage{booktabs}
\title{HiVAE: Hierarchical Latent Variables \\ for Scalable Theory of Mind}
\author {
    Nigel Doering\textsuperscript{\rm 1},
    Rahath Malladi\textsuperscript{\rm 1},
    Arshia Sangwan\textsuperscript{\rm 2}, 
    David Danks\textsuperscript{\rm 3},
    Tauhidur Rahman\textsuperscript{\rm 1}
}
\affiliations{
    \textsuperscript{\rm 1}University of California San Diego, School of Computing, Information, and Data Sciences\\
    \textsuperscript{\rm 2}New York University\\
    \textsuperscript{\rm 3}University of Virginia}

\begin{document}
\maketitle

\begin{abstract}
Theory of mind (ToM) enables AI systems to infer agents' hidden goals and mental states, but existing approaches focus mainly on small human understandable gridworld spaces. We introduce HiVAE, a hierarchical variational architecture that scales ToM reasoning to realistic spatiotemporal domains. Inspired by the belief-desire-intention structure of human cognition, our three-level VAE hierarchy achieves substantial performance improvements on a 3,185-node campus navigation task. However, we identify a critical limitation: while our hierarchical structure improves prediction, learned latent representations lack explicit grounding to actual mental states. We propose self-supervised alignment strategies and present this work to solicit community feedback on grounding approaches.
\end{abstract}

\section{Introduction}
While Theory of Mind (ToM) has driven advances in AI for human-interpretable tasks \cite{baker2009action, baker2017rational, rabinowitz2018machine, jin2024mmtom, zhang2025autotom}, existing architectures remain largely untested on large-scale trajectory datasets such as pedestrian paths, ship tracking, or wearable sensor data.

\begin{figure*}[h]
    \centering
    \includegraphics[width=\linewidth]{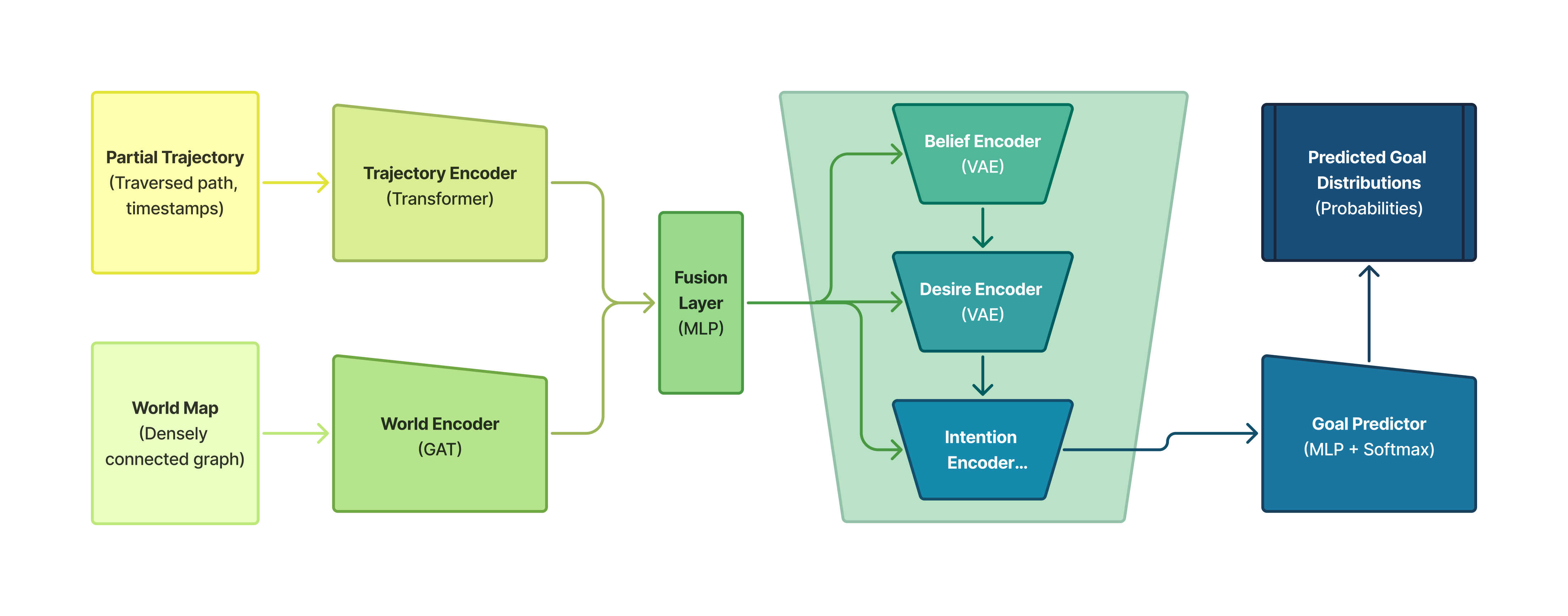}
    \caption{HiVAE first encodes the agent’s partial trajectory and the environment graph into a unified latent representation. This representation feeds a hierarchical mind-state module, sequentially inferring beliefs, desires, and intentions, which then drives the goal predictor to output a probability distribution over all possible goals.}
    \label{fig:architecture}
\end{figure*}

We introduce HiVAE, a hierarchical variational architecture motivated by the belief-desire-intention (BDI) structure \cite{rao1995bdi}, where beliefs inform desires, which inform intentions (Figure \ref{fig:architecture}). Our three-level VAE hierarchy \cite{kingma2013auto} operating on trajectory and graph encodings achieves substantial performance gains on a 3,185-node campus navigation task. However, while this hierarchical structure improves prediction, the learned latent representations lack explicit grounding to actual mental states-a key limitation we address through proposed self-supervised alignment strategies.

\section{Architecture}

Our approach models goal inference as a hierarchical latent variable problem inspired by the belief-desire-intention (BDI) structure of human mental states \cite{rao1995bdi}. Given an agent's partial trajectory $\mathcal{T} = \{v_t\}_{t=1}^{T}$ where $v_t \in \mathcal{V}$ represents visited nodes at times $t$, and a spatial graph $\mathcal{G} = (\mathcal{V}, \mathcal{E})$ representing the environment, we predict the agent's goal $g \in \mathcal{V}$ through three main components.

\textbf{Trajectory-Graph Encoder.} We process both temporal movement patterns and spatial structure through parallel encoding pathways. The trajectory encoder applies a Transformer architecture with temporal positional encodings to capture movement dynamics:
$$\mathbf{h}_{\text{traj}} = \frac{1}{T}\sum_{t=1}^T \text{Transformer}(\text{Embed}(v_t) + \text{TimeEmbed}(t))_t$$
where $\text{Embed}(\cdot)$ maps nodes to learned embeddings and $\text{TimeEmbed}(\cdot)$ uses sinusoidal encoding for temporal patterns.

The graph encoder uses Graph Attention Networks \cite{GAT} to capture spatial relationships. For each node $i$, attention weights over neighbors $j \in \mathcal{N}_i$ enable the model to weight spatial proximity and connectivity:
$$\alpha_{ij} = \frac{\exp(\text{LeakyReLU}(\mathbf{a}^T[\mathbf{W}\mathbf{h}_i \| \mathbf{W}\mathbf{h}_j]))}{\sum_{k \in \mathcal{N}_i} \exp(\text{LeakyReLU}(\mathbf{a}^T[\mathbf{W}\mathbf{h}_i \| \mathbf{W}\mathbf{h}_k]))}$$

We fuse encodings via: $\mathbf{h}_{\text{fused}} = \text{MLP}([\mathbf{h}_{\text{traj}} \| \mathbf{h}_{\text{graph}}])$, creating a unified spatiotemporal representation.

\textbf{Hierarchical Mind-State VAE.} Inspired by cognitive models where beliefs inform desires, which inform intentions, we construct a three-level hierarchical VAE. Each level $l \in \{b, d, i\}$ learns a latent variable conditioned on the fused encoding and all previous mental states:
\begin{align*}
q_\phi(\mathbf{z}_l | \mathbf{h}_{\text{fused}}, \mathbf{z}_{<l}) &= \mathcal{N}(\mathbf{z}_l; \boldsymbol{\mu}_l, \text{diag}(\boldsymbol{\sigma}_l^2)) \\
\boldsymbol{\mu}_l, \log\boldsymbol{\sigma}_l^2 &= f_\phi([\mathbf{h}_{\text{fused}} \| \mathbf{z}_{<l}])
\end{align*}
where $\mathbf{z}_{<l}$ denotes concatenation of previous levels (e.g., $\mathbf{z}_{<d} = \mathbf{z}_b$, $\mathbf{z}_{<i} = [\mathbf{z}_b \| \mathbf{z}_d]$), and $f_\phi$ is a shared encoder network. Each level includes reconstruction objectives to encourage meaningful representations, though these are not explicitly supervised to correspond to actual mental states-a key limitation we address in future work.

\textbf{Goal Prediction and Training.} We map concatenated mental states to goal probabilities:
$p_\theta(g|\mathbf{z}_b, \mathbf{z}_d, \mathbf{z}_i) = \text{Softmax}(f_\theta([\mathbf{z}_b \| \mathbf{z}_d \| \mathbf{z}_i]))$

Training minimizes a composite objective:
$\mathcal{L}_{\text{total}} = \mathcal{L}_{\text{goal}} + \beta_{\text{KL}} \mathcal{L}_{\text{KL}} + \beta_{\text{recon}} \mathcal{L}_{\text{recon}}$
where $\mathcal{L}_{\text{goal}}$ is cross-entropy loss over goals, $\mathcal{L}_{\text{KL}}$ regularizes the latent space with a hierarchical prior, and $\mathcal{L}_{\text{recon}}$ ensures each VAE level learns meaningful representations.

\section{Preliminary Results}

We evaluate on synthetic pedestrian trajectories over a real-world campus graph (3,185 nodes, 9,000 edges from OpenStreetMap). We simulate 100 agents with individualized preferences, each generating 1,000 episodes via shortest-path planning, yielding 100K trajectories split 70/30 train/test.

Figure 2 shows Brier scores at varying observation fractions. HiVAE  consistently achieves the lowest Brier scores across all path fractions, outperforming all baselines by a large margin. A Wilcoxon Signed Rank test comparing HiVAE to the next highest performer confirms the significance of the improvement $(p<0.01)$. Two robustness tests demonstrate: (1) when agents pass near least-preferred goals, our model maintains low false-goal probability while distance-based baselines are misled; (2) under significant preference drift, our model generalizes better to non-stationary behavior. An ablation study confirms the full three-level hierarchy (BDI) outperforms two-level (BD) and single-level (B) variants. See appendix section Additional Experiments  and section Additional Results for further details.

\begin{figure}[h]
    \centering
    \includegraphics[width=\linewidth]{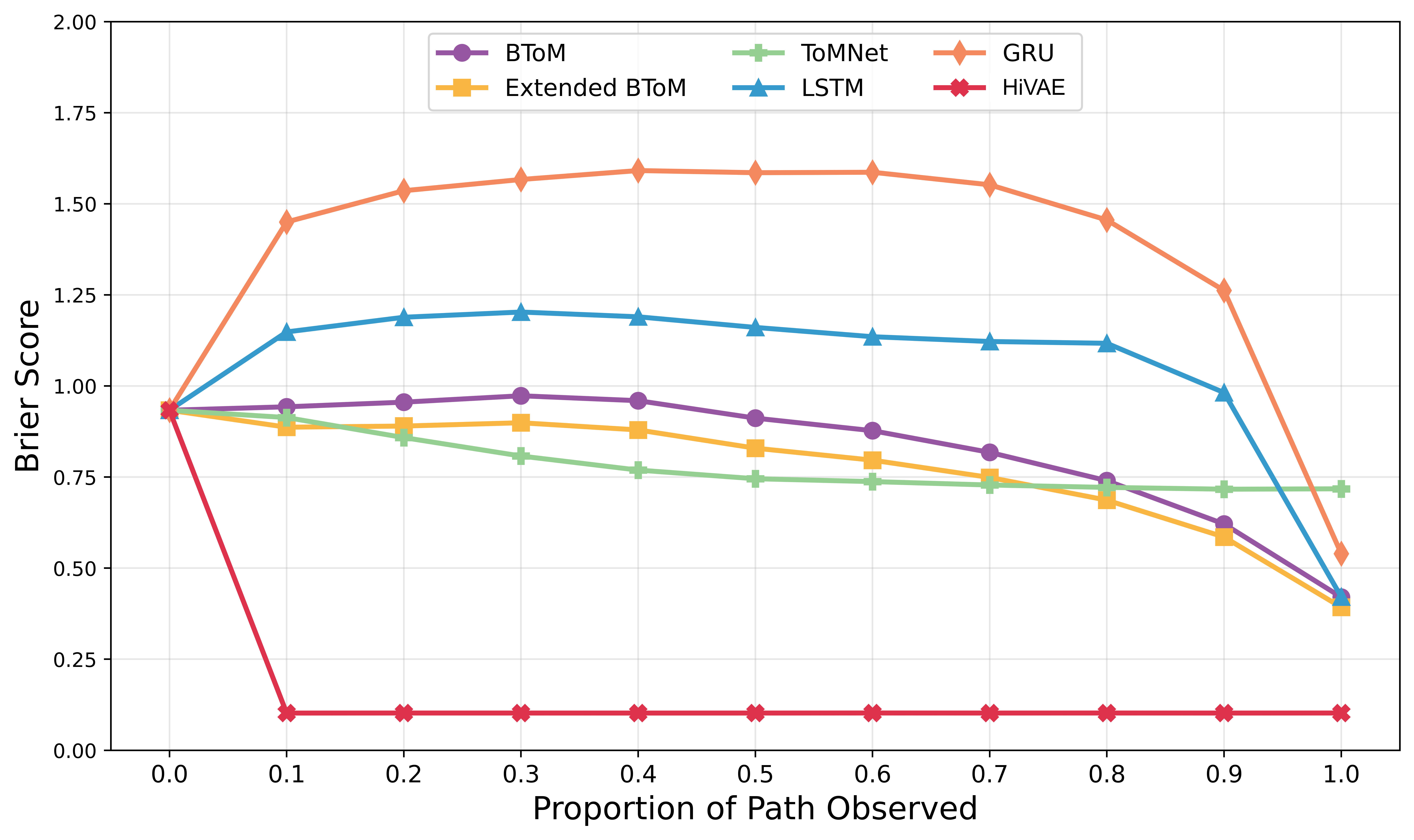}
    \caption{Overall performance of models on goal prediction averaged across all trajectories. Lower is better.}
    \label{fig:exp_1-overall_brier_results}
\end{figure}

\section{Limitations \& Future Work}

\textbf{Grounding Mental States.} The latent variable layers are supervised only through reconstruction and goal prediction. While our ablation study confirms this hierarchical structure improves performance, the learned representations lack explicit grounding to actual mental states. We propose self-supervised objectives to align latent layers with ToM concepts: (1) contrastive learning where trajectories sharing goals have similar desire embeddings $\mathbf{z}_d$; (2) auxiliary prediction where beliefs $\mathbf{z}_b$ predict observable environment features; (3) intention consistency where $\mathbf{z}_i$ aligns with short-term waypoint predictions.

\textbf{Agent Personalization.} The current architecture treats all agents identically. Following ToMNet \cite{rabinowitz2018machine}, we will incorporate meta-learned agent embeddings to capture individual preferences while maintaining population-level patterns, enabling improved personalization in goal inference.

\textbf{Architectural Alternatives.} While we employ hierarchical VAEs for their probabilistic structure, alternative latent architectures such as normalizing flows \cite{rezende2015variational}, vector quantization \cite{van2017neural}, or diffusion models \cite{ho2020denoising} may offer advantages for mental state representation and merit future exploration.

\textbf{Real-World Deployment.} Future work will bridge theoretical ToM models with real-world trajectory data from wearables, mobile sensing, and non-spatial domains such as healthcare monitoring.

\bibliography{aaai2026}

\appendix
\section{Appendix}

\section{Pedestrian Simulation}
To evaluate our approach in a structured yet realistic setting, we created a synthetic pedestrian dataset grounded in real-world geography. Our environment supports complex agent behavior while preserving full control over latent variables such as preferences and goals.

We constructed a directed graph of 3,185 nodes and 9,000 edges from OpenStreetMap\cite{openstreetmap}, representing walkable campus paths and intersections. Points of interest-academic buildings, dining halls, dorms-were mapped to nodes and designated as candidate goals.

Each of the 100 agents was assigned a unique goal preference distribution, drawn from a Dirichlet prior to reflect habitual patterns. We simulated 1,000 episodes per agent by sampling origin-goal pairs and computing shortest paths, yielding 100,000 trajectories.

The dataset was split 70/30 per agent into training and test sets. Each trajectory includes agent ID, episode ID, timestamps, and full path data, supporting partial trajectory inference during evaluation.

A key limitation in this generated dataset was the use of only shortest path routes. We fix this in the next implementation of the simulation by drawing from the top 5 shortest paths using a categorical probability distribution based on the length of the path. So shorter paths are more probable. This is intended to reflect expected utility for the agent.

\section{Additional Experiments} 
\label{section:additional_exps}

\subsection{Experiment 1: Goal Inference Over Time}
This experiment measures how quickly and accurately models infer intent from partial movement. Using the 30\% held‐out test trajectories, we evaluate at 25\%, 50\%, 75\%, and 95\% observation fractions. At each prefix, models predict the destination over the full 3,185‐node goal set.

This setting presents a challenging inference task: the environment is a large, densely connected graph with many plausible goals, and early trajectory segments may align with multiple destinations. Unlike prior ToM benchmarks set in simplified gridworlds, our setting requires models to disambiguate intent under greater spatial and behavioral ambiguity.

We measure performance using the Brier Score, a proper scoring rule that evaluates the accuracy and calibration of probabilistic predictions:
\begin{align*}
\mathrm{Brier}(p,y) &= \sum_{i=1}^{N} (p_i - y_i)^2
\end{align*}
where $p_i$ is the predicted probability of goal $i$, and $y_i$ is the one‐hot ground truth. We report mean Brier Scores and visualize model performance trends to highlight each model’s inference dynamics.

\subsection{Experiment 2: False Goal Test}

Our second experiment probes whether models can distinguish an agent’s true intent when presented with a salient but misleading alternative. This setup evaluates whether models rely solely on spatial proximity or whether they internalize agents’ long-term preferences.

For each agent \(i\), we identify their most and least preferred destinations from their goal distribution \(G_i\):
\[
g^*_i = \arg\max_{g \in G_i} P(g), \quad \tilde{g}_i = \arg\min_{g \in G_i} P(g),
\]
where \(g^*_i\) is the agent’s true goal and \(\tilde{g}_i\) serves as the false goal.

We generate a synthetic trajectory in which the agent must pass near \(\tilde{g}_i\) en route to \(g^*_i\). This creates an ambiguous context that tests whether models can resist misleading proximity cues and instead leverage learned priors over agent behavior.

We measure the \textbf{False Goal Probability}-the model’s predicted probability for \(\tilde{g}_i\) at each timestep-as the primary evaluation metric. This is aggregated across agents, using one false-goal episode per agent. The resulting trend reveals whether each model is susceptible to distraction from nearby but implausible goals, or whether it correctly integrates behavioral history into its inferences.

\subsection{Experiment 3: Generalizability}
Our third experiment investigates how well each model generalizes when an agent’s preferences-i.e., their goal distribution---change over time. This scenario tests the robustness of models to non-stationary behavior and distributional shift, which commonly occurs in real-world agents (e.g., humans changing routines).

For each agent $i$, let $G_i$ denote their original categorical distribution over goals. We generate a perturbed distribution $G_i'$ such that the Kullback–Leibler divergence satisfies:
\[
\text{KL}(G_i \,||\, G_i') > 1
\]
This threshold ensures a meaningful shift in the agent’s preferences, rather than minor noise. We then sample new goal episodes from $G_i'$ to construct a new test dataset of 100 episodes per agent.

Each model is evaluated on this drifted dataset using the same metric from Experiment 1: \textbf{Brier score} at various observation percentages (25\%, 50\%, 75\%, 95\%). Models like BToM, which infer goals purely from current trajectories without learning historical preferences, are expected to be unaffected. In contrast, models that rely on learned priors may exhibit a drop in performance, thereby highlighting trade-offs between generalization and adaptability.

\section{Additional Results}
\label{section:additional_results}

In our first experiment, we evaluate the goal prediction performance of all baseline and proposed models using the Brier score at different path completion percentages. Brier scores are computed at each 10\% interval of the trajectory, and then averaged across all trajectories in the validation set. Table~\ref{tab:pedestrian-brier} and Figure~\ref{fig:exp_1-overall_brier_results} summarize the results at key checkpoints (25\%, 50\%, 75\%, and 95\%).

Our proposed model, HiVAE, consistently achieves the lowest Brier scores across all path fractions, outperforming all baselines by a large margin. Notably, while traditional models such as BToM and Extended BToM improve slightly over time, their overall prediction accuracy remains substantially lower than HiVAE, which exhibits stable and highly accurate predictions even with minimal trajectory input. GRU and LSTM models fail to converge to accurate distributions, exhibiting erratic Brier scores throughout.

To statistically validate the performance gap, we conduct a Wilcoxon signed-rank test comparing HiVAE against the next-best model, Extended BToM, across ten held-out validation trajectories. The result \((W = 0,\; Z = -2.80,\; p < 0.01)\) indicates that HiVAE significantly outperforms Extended BToM in all ten cases, confirming the robustness of our model’s predictions.

\begin{table}[ht]
\centering
\small
\caption{Brier Score on Pedestrian Dataset at Different Path Completion Percentages - Experiment 1. Best results shown in bold.}
\label{tab:pedestrian-brier}
\begin{tabular}{lcccc}
\toprule
\textbf{Model} & 25\% & 50\% & 75\% & 95\% \\
\midrule
Bayesian ToM (BToM)  & 0.9555 & 0.9117 & 0.7400 & 0.4204 \\
Extended BToM     & 0.8899 & 0.8292 & 0.6865 & 0.3927 \\
GRU     & 1.5358 & 1.5853 & 1.4555 & 0.5399 \\
LSTM     & 1.1886 & 1.1606 & 1.1171 & 0.4211 \\
ToMNet         & 0.8583 & 0.7453 & 0.7215 & 0.7176 \\
HiVAE (Ours) & \textbf{0.1042} & \textbf{0.1036} & \textbf{0.1023} & \textbf{0.1023} \\
\bottomrule
\end{tabular}
\end{table}

\begin{figure}[h]
    \centering
    \includegraphics[width=\linewidth]{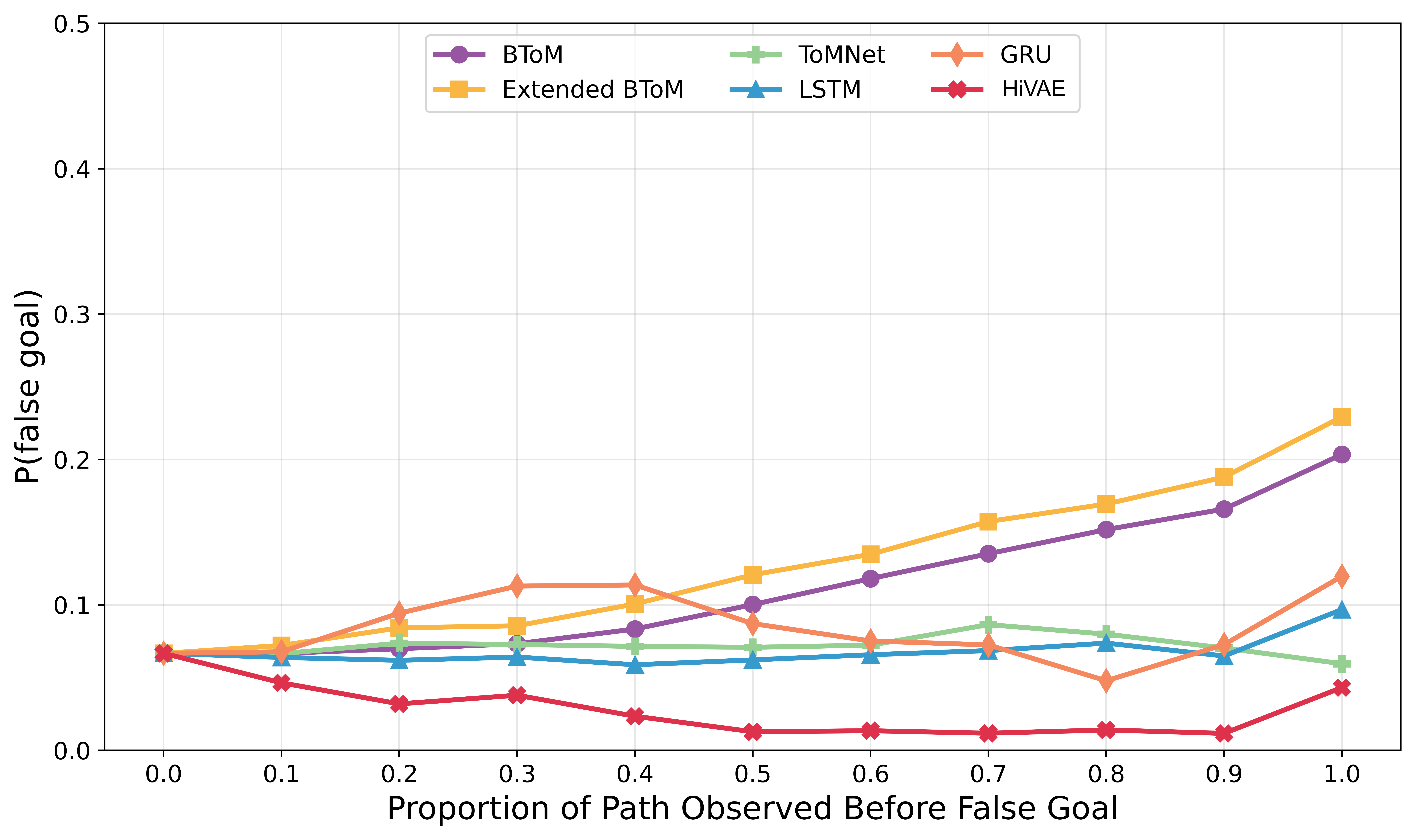}
    \caption{Probability of False Goal on Pedestrian Dataset at Different Path Completion Percentages Leading up to the False Goal - Experiment 2. Lower is better.}
    \label{fig:exp_2-false_goal}
\end{figure}

Figure~\ref{fig:exp_2-false_goal} presents the results of Experiment 2, where we evaluate each model's susceptibility to misleading false goals. Specifically, we plot the probability assigned to the false goal across 10 intervals leading up to the point at which the agent passes near the false goal location.

The results demonstrate a clear advantage for HiVAE. Unlike other models, which steadily increase their belief in the false goal as the agent approaches it, HiVAE consistently assigns a low probability to this distractor. This suggests that HiVAE successfully integrates historical preference information, allowing it to resist spurious inferences driven by short-term proximity cues. In contrast, learning-based baselines such as GRU, LSTM, and ToMNet show moderate increases in false goal probability as the agent nears the distractor. Notably, this experiment also highlights a key vulnerability of purely distance-based Bayesian models, which are easily misled as the agent moves closer to the false goal.

Overall, these findings indicate that HiVAE not only improves goal inference accuracy over time, but also maintains robustness in the presence of conflicting evidence, highlighting its ability to reason about intent rather than mere motion.

\begin{table}[ht]
\centering
\small
\caption{Difference in Brier Score Between New and Original Pedestrian Dataset at Different Path Completion Percentages - Experiment 3. Closer to zero indicates same performance. Negative indicates model is performing better on new data.}
\label{tab:ais-brier}
\begin{tabular}{lcccc}
\toprule
\textbf{Model} & 25\% & 50\% & 75\% & 95\% \\
\midrule
Bayesian ToM (BToM)     & 0.0644 & 0.0532 & 0.3755 & 0.8350 \\
Extended BToM     & 0.0904 & 0.2814 & 0.5871 & 1.0222 \\
GRU     & -0.0184 & 0.0170 & 0.0290 & 0.6185 \\
LSTM     & 0.0104 & -0.0071 & 0.0373 & 0.2436 \\
ToMNet           & -0.0182 & -0.0342 & -0.0343 & -0.0275 \\
HiVAE (Ours) & 0.0205 & 0.0309 & 0.0307 & 0.0307 \\
\bottomrule
\end{tabular}
\end{table}

\begin{figure}
    \centering
    \includegraphics[width=\linewidth]{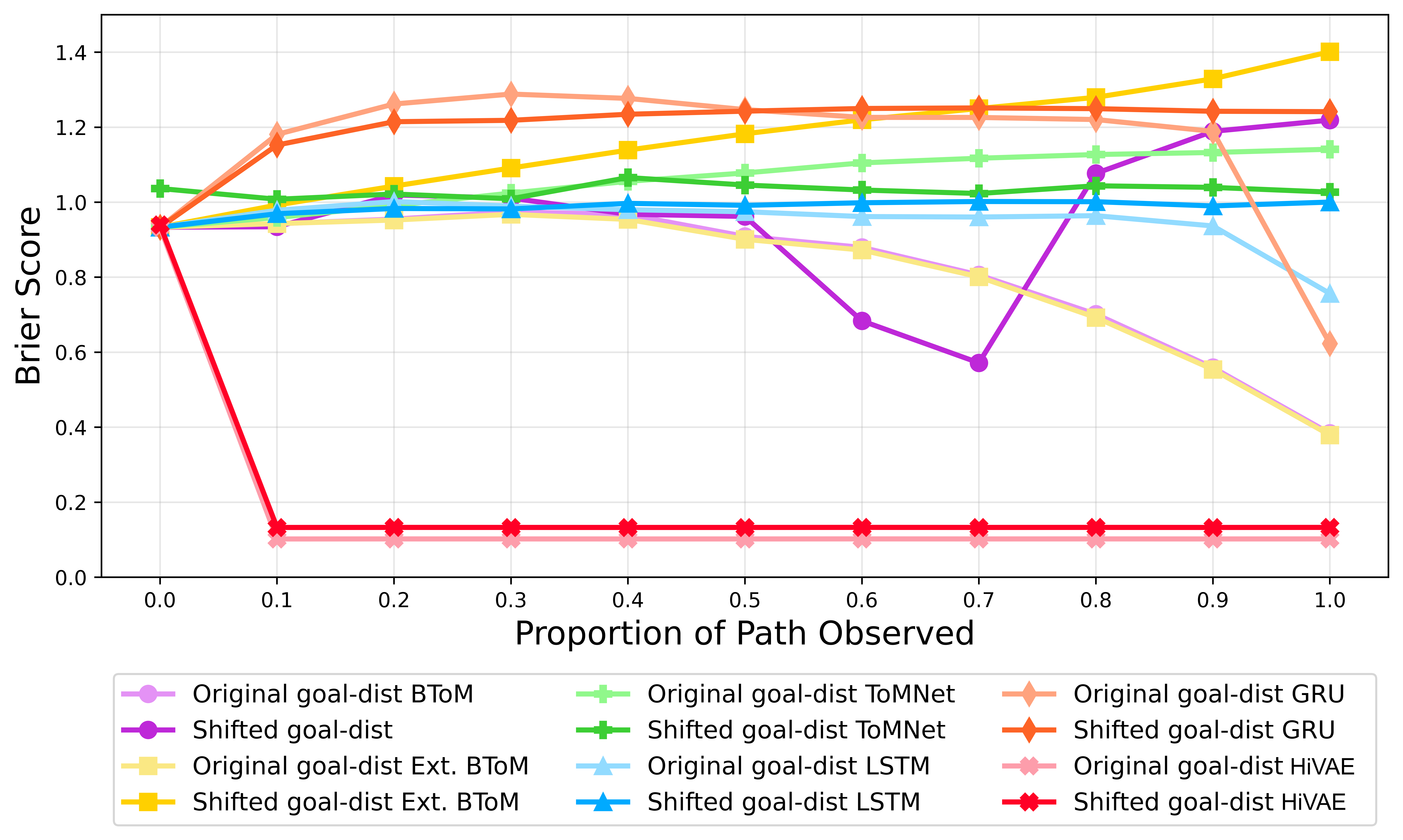}
    \caption{Overall performance of models on goal prediction
averaged across all trajectories for both original and new pedestrian datasets. Lower is better.}
    \label{fig:exp_3}
\end{figure}

Our third experiment tests each model’s ability to generalize under distributional shift, specifically, when an agent’s goal preferences change significantly over time. This scenario reflects real-world non-stationarity, such as changes in human routines, and probes whether models overly rely on static prior knowledge.

Figure~\ref{fig:exp_3} shows the Brier scores of various models on the original and the drifted test sets, and Table~\ref {tab:ais-brier} refers to the change in Brier Score between these test sets. Values closer to zero indicate better robustness to preference shift, while larger positive differences suggest a model is overfitting to historical preferences. Across all observation levels, models that heavily leverage learned priors, especially Extended BToM, show the largest shifts in Brier score (up to –1.02 at 95\%), indicating they respond most strongly (for better or worse) when goals drift. In contrast, ToMNet’s differences remain near zero, and our HiVAE consistently hovers around –0.03, demonstrating the greatest stability (and mild improvement) under non-stationary preference shifts.

\end{document}